\documentclass[letterpaper]{article} 
\usepackage{aaai25}
\usepackage{times}  
\usepackage{helvet}  
\usepackage{courier}  
\usepackage[hyphens]{url}  
\usepackage{graphicx} 
\usepackage{natbib}  
\usepackage{caption} 
\usepackage{algorithm}
\usepackage{algorithmic}

\usepackage{amssymb}
\usepackage{amsmath}
\usepackage{colortbl} 
\usepackage{array} 
\usepackage{booktabs} 
\usepackage{multirow} 
\usepackage{makecell}

\usepackage{newfloat}
\usepackage{listings}
\DeclareCaptionStyle{ruled}{labelfont=normalfont,labelsep=colon,strut=off} 
\floatstyle{ruled}
\newfloat{listing}{tb}{lst}{}
\floatname{listing}{Listing}
\title{sTransformer: A Modular Approach for Extracting Inter-Sequential and Temporal Information for Time-Series Forecasting}

\author {
    Jiaheng Yin \textsuperscript{\rm 1},
    Zhengxin Shi\textsuperscript{\rm 2},
    Jianshen Zhang \textsuperscript{\rm 2},
    Xiaomin Lin \textsuperscript{\rm 2},
    Yulin Huang \textsuperscript{\rm 2},\\
    Yongzhi Qi  \textsuperscript{\rm 2} $^*$,
    Wei Qi \textsuperscript{\rm 1} \thanks{Corresponding author.}
}
\affiliations {
    \textsuperscript{\rm 1} Tsinghua University, Beijing, China\\
    \textsuperscript{\rm 2} JD.com, Beijing, China\\
    yjh22@mails.tsinghua.edu.cn, 
    \{shizhengxin1, zhangjianshen, linxiaoming7, huangyulin16, qiyongzhi1\}@jd.com, 
    qiw@tsinghua.edu.cn
}
\copyrighttext{}

\begin{document}

\maketitle

\begin{abstract}
In recent years, numerous Transformer-based models have been applied to long-term time-series forecasting (LTSF) tasks. However, recent studies with linear models have questioned their effectiveness, demonstrating that simple linear layers can outperform sophisticated Transformer-based models. 
In this work, we review and categorize existing Transformer-based models into two main types: (1) modifications to the model structure and (2) modifications to the input data. The former offers scalability but falls short in capturing inter-sequential information, while the latter preprocesses time-series data but is challenging to use as a scalable module. 
We propose $\textbf{sTransformer}$, which introduces the Sequence and Temporal Convolutional Network (STCN) to fully capture both sequential and temporal information. Additionally, we introduce a Sequence-guided Mask Attention mechanism to capture global feature information. Our approach ensures the capture of inter-sequential information while maintaining module scalability. 
We compare our model with linear models and existing forecasting models on long-term time-series forecasting, achieving new state-of-the-art results. We also conducted experiments on other time-series tasks, achieving strong performance. These demonstrate that Transformer-based structures remain effective and our model can serve as a viable baseline for time-series tasks.

\end{abstract}

%

\section{Introduction}

Transformer \cite{vaswani2017attention} architecture has achieved great success in various fields, such as natural language processing (NLP) \cite{kalyan2021ammus,gillioz2020overview}, computer vision (CV) \cite{liu2023survey, wu2021cvt, dosovitskiy2020image}, and speech \cite{karita2019comparative,huang2020conv}. In the field of time-series forecasting, its attention mechanism can automatically learn the connections between elements in a sequence, leading to widespread application \cite{lim2021temporal,wen2022transformers}. Informer \cite{zhou2021informer}, Autoformer \cite{wu2021autoformer}, and FEDformer \cite{zhou2022fedformer} are successful Transformer variants applied in time-series forecasting.

\begin{figure}[ht]
\centering
\includegraphics[width=0.8\columnwidth]{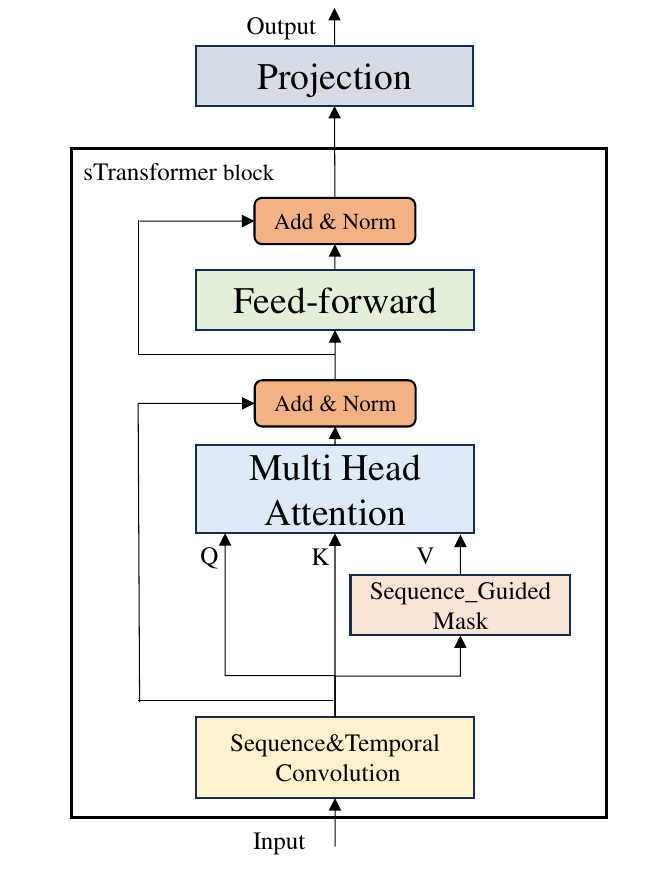} 
\caption{sTransformer block overview. STCN and SeqMask are introduced into the traditional Transformer structure. STCN extracts information from both sequence and temporal aspects. SeqMask interacts features of the Value layer with global features, enhancing global representation capability.}
\label{Overall}
\end{figure}

Recent research \cite{zeng2023transformers} has shown that simple linear structures have outperformed previous models, challenging the effectiveness of the Transformer architecture in time-series forecasting. In response to this criticism, new paradigms have been proposed, such as iTransformer \cite{liu2023itransformer} and PatchTST \cite{nie2022time}. 
They demonstrate that previous models were an inappropriate use of the Transformer structure. iTransformer embeds each practice sequence into variate tokens, allowing the attention mechanism to capture multivariable correlations. PatchTST constructs novel patches, transforming the original sequence into multiple subsequences to enhance local contextual information capture. These models indicate that the Transformer structure is still effective in time-series forecasting, but the key lies in enabling the model to capture more additional information about sequences, thereby improving its representation capacity. Furthermore, most current improvements focus on modifying data input rather than making significant changes to the components of the Transformer.

Based on capturing information between sequences and the modularization of Transformer components, we propose a new paradigm called \textbf{sTransformer}. Within the Transformer structure, we introduce two components: Sequence and Temporal Convolutional Network (\textbf{STCN}) and Sequence-guided Mask Attention (\textbf{SeqMask}). STCN extracts information from both the inter-sequential and temporal dimensions, allowing it to focus on relationships across different time steps and the influence of multiple variables. SeqMask enables value in attention to consider more global information. These components significantly enhance the representation capacity of the Transformer. We demonstrate the superiority of our model on several commonly used public datasets, surpassing the linear DLinear model and outperforming the latest state-of-the-art models, establishing a new SOTA for long-term time-series forecasting.

Our work contributes as follows:
\begin{itemize}
    \item We constructed the STCN network structure, which uses temporal convolution to capture temporal correlations across different time steps, and inter-sequential convolution to capture correlations between sequences, thereby enhancing the representation capability of attention inputs.
    \item We developed the Sequence-guided Mask attention mechanism, enabling the value layer to perform feature interactions and acquire global information.
    \item We designed the highly scalable sTransformer block, integrating the STCN and SeqMask mechanisms into the Transformer structure. Multiple layers of blocks can be embedded in the framework to enhance the extraction of features from sequential and temporal dimensions.
\end{itemize}

\section{Related Works}
\subsection{Transformer-based Long-term Time-Series Forecasting}
Numerous recent studies have applied Transformer structure to long-term time-series forecasting tasks. These works can be categorized into two types: (1) modifications to the model structure and (2) modifications to the input data. In Table \ref{tab: Transformer models}, we present some of the major existing research works and compare their advantages and disadvantages.


Models with updated components include Autoformer \cite{wu2021autoformer}, Informer \cite{zhou2021informer}, FEDformer \cite{zhou2022fedformer}, Crossformer \cite{zhang2023crossformer}. These models mainly focus on the attention mechanism's modeling of the temporal dimension and the improvement of complexity for long sequences. However, with the emergence of linear predictors \cite{oreshkin2019n, zeng2023transformers, das2023long}, models with updated components have shown inferior performance compared to linear predictors.
Therefore, approaches with modification to the time-series inputs emerge \cite{liu2022non,nie2022time,liu2023itransformer}. These models focus on the input data structure, directly or through construction, extracting the correlation information within and between sequences.
We believe the relatively poor performance of the first approach is not due to the component updates but rather due to the weak ability to extract correlation information between sequences. While the second approach extracts information intuitively, it has poor scalability. We believe that by designing components that can effectively extract inter-sequence correlations, we can achieve better scalability and surpass the predictive performance of the second type of method.

\begin{table*}[ht] 
\centering
\begin{tabular}{c|m{6.5cm}|m{6.5cm}}
\toprule[2pt]
\textbf{Type}
& \multicolumn{1}{c|}{\textbf{Modification to the Structure}} 
& \multicolumn{1}{c}{\textbf{Modification to the Input Data}}  \\
\midrule[2pt]
\textbf{Interpretation}
& These models adjust the Transformer's internal components to enable the attention module to model the temporal dimension and extract complex information from long sequences.
& These models mainly focus on altering the structure of input data, allowing Transformer to capture temporal features more directly.\\
\midrule
\multicolumn{1}{c|}{\makecell{\textbf{Representative} \\ \textbf{Models}} } 
& 
\textbf{LogSparse}:
proposes convolutional self-attention, generates queries and keys through \textit{causal convolution}, enabling the attention mechanism to capture local context information better while reducing memory cost.

\textbf{Autoformer}:
performs-series decomposition and introduces an \textit{auto-correlation} mechanism for aggregating temporal information.

\textbf{Other works}: 
\textbf{Informer}, \textbf{FEDformer}, $\dots$
& 
\textbf{PatchTST}:
constructs patches to divide the time-series into multiple sub-sequences, enhancing the capture of local contextual information.

\textbf{iTransformer}:
extracts each time point of the time-series into variate tokens, capturing the correlations between multiple variables in an "inverted" manner.
\\
\midrule
\multicolumn{1}{c|}{\textbf{Characteristics}}  
&
\textbf{Advantages}:

Enhanced scalability

\textbf{Disadvantages}:

(1) Ignoring sequence correlation: Focusing only on temporal information. 

(2) Inferior to linear models: Simple linear model (DLinear) outperforms transformer-based models with updated structure on common datasets and metrics.
& 
\textbf{Advantages}:

(1) Sequence correlation: Information between sequences can be captured.

(2) Superior to linear model (DLinear).

\textbf{Disadvantages}:

Limited scalability.
\\
\bottomrule[2pt]
\end{tabular}%
\caption{Comparison of two types of Transformer-based time-series forecasting models.}
\label{tab: Transformer models}
\end{table*}

\subsection{CNN in Time-Series Forecasting}



The Transformer architecture excels at handling long-range dependencies, while CNNs are very effective at capturing local features. In recent years, some research has combined CNNs with the Transformer architecture to leverage the strengths of both, applying them to time-series problems.
Transformer models combined with CNN primarily utilize the concept of convolution to capture local information across time steps. 
The introduction of Temporal Convolutional Networks (TCN) \cite{bai2018empirical} architecture enhances the memory capacity for long sequences, which has led to its application in time-series task \cite{franceschi2019unsupervised}.
LogSparse \cite{li2019enhancing} uses convolutional kernels with a stride greater than 1 when computing Query and Key, enabling the attention mechanism to focus on contextual information in the temporal dimension. Related models mainly capture local information in the temporal domain, which weakens the feature extraction capability of CNN and is the reason for the limited improvement of CNN-based Transformer. We extend the concept of convolution to inter-sequence relations, simultaneously capturing relevant information from both the temporal and inter-sequential dimensions.

\subsection{Instance-Guided Mask}

MaskNet \cite{wang2021masknet} is proposed to improve Click-Through Rate (CTR) estimation. They construct an instance-guided mask method, which performs an element-wise product between feature embedding and input instance-guided feed-forward layers in DNN. This method integrates global information into the embedding and feed-forward layers through the mask.
There are also methods that use feature interaction to extract global information \cite{wang2022enhancing}. These methods have been applied in the recommendation field but, to our knowledge, have not been applied to time-series forecasting and Transformer modification. Each time-series can also be considered as an instance, so we propose a similar concept called sequence-guided mask to assist Transformer in extracting more global contextual information.


\section{sTransformer}
The time-series forecasting problem can be defined as: given a historical dataset of $M$ sequences (variables), where one sequence $i\in \{1,2,\dots, M\}$ corresponds to the time-series $(x_{i,1},x_{i,2}, \dots, x_{i, T})$, and we aim to predict the output for the next $K$ time periods $(x_{i, T+1},x_{i, T+2},\dots,x_{i, T+K})$. Here we use $\mathbf{x}_{:, 1:T} \in \mathbb{R}^{M\times T}$ to denote the concatenation of $M$ time-series from $1$ to $T$, and $ \mathbf{x}_{:,T+1:T+K} \in \mathbb{R}^{M\times K}$ to denote the concatenation from $T+1$ to $T+K$.

\subsection{Structure overview}

\subsection{STCN}

\begin{figure*}[ht]
\centering
\includegraphics[width=1.95\columnwidth]{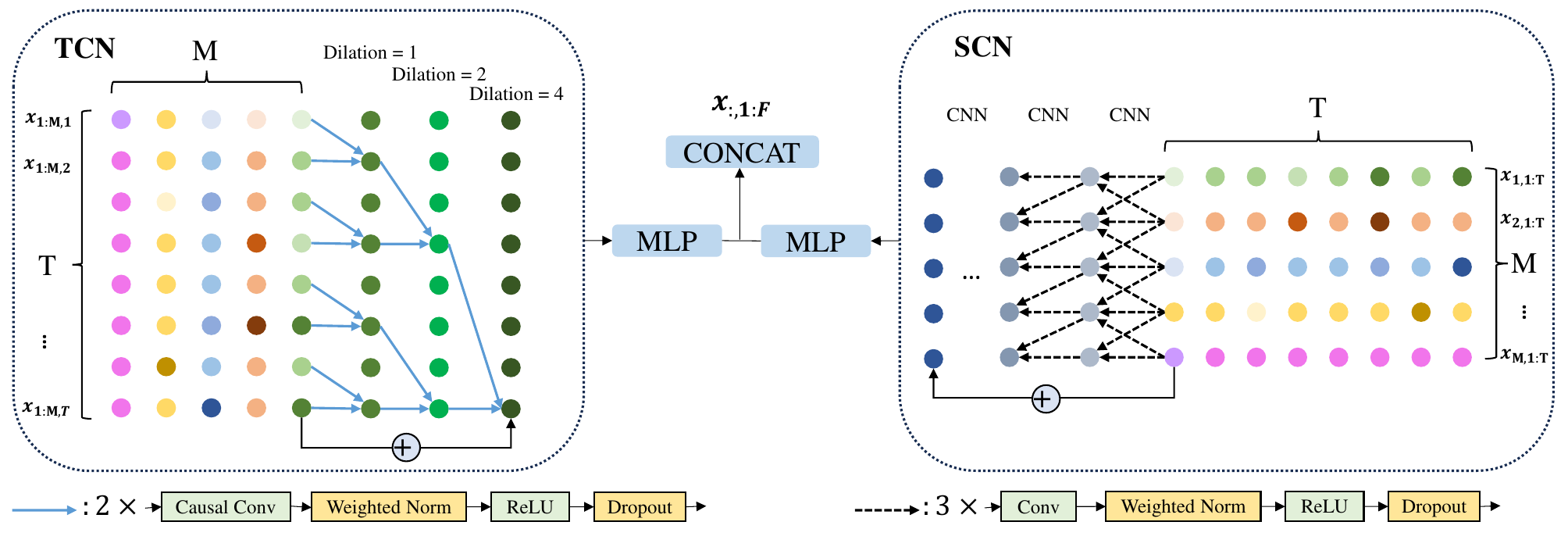}
\caption{STCN. The left part is the TCN structure, and the right part is the SCN structure. TCN performs convolution along the temporal dimension, receiving information from previous time steps at each position of each dilation layer. SCN performs convolution along the sequence dimension, using padding through concatenation. In TCN, layers employ different value of dilation, while in SCN, layers use varying convolution kernel sizes. 
In each layer of TCN and SCN, two sets and three sets of convolutional blocks are integrated respectively. 
Notably, due to the temporal property, the convolutions in TCN are causal.}
\label{STCN}
\end{figure*}

The information in time-series data is manifested at two levels: the sequence level and the temporal level. We designed a Sequence and Temporal Convolutional Network (STCN) to extract information from both levels simultaneously. The STCN maps the temporal feature space to a new feature space $\text{STCN}(\cdot): \mathbb{R}^{M\times T} \rightarrow \mathbb{R}^{M\times F}$, enabling each sequence to focus on its own temporal information while also capturing shared information across other sequences. Figure \ref{STCN} shows the complete structure of STCN.

\begin{equation}
    \mathbf{x}_{:,1:F} = \text{STCN}(\mathbf{x}_{:,1:T}) 
\end{equation}

\subsubsection{Temporal convolution.} 
We first apply TCN for temporal convolution on the raw data, then we use an MLP to extract temporal-level information. 

\begin{equation}
\begin{aligned}
    &\mathbf{x}_{:,1:T}^{(tcn)} = \text{TCN}(\mathbf{x}_{:,1:T}) \in \mathbb{R}^{M\times T}, \\
    &\mathbf{x}_{:,1:\frac F2}^{mlp(1)} = \text{MLP}^{(1)}(\mathbf{x}_{:,1:T}^{(tcn)}) \in \mathbb{R}^{M\times \frac{F}{2}}.
\end{aligned}
\end{equation}

\subsubsection{Sequence convolution.}
Similar to temporal convolution, we use SCN for convolution across sequences, followed by another MLP to extract inter-sequence information. Here, $a'$ denotes the transpose of $a$.

\begin{equation}
\begin{aligned}
    &\mathbf{x}_{:,1:T}^{(scn)} = \text{SCN}(\mathbf{x}_{:,1:T}') \in \mathbb{R}^{d_s \times M}, \\
    &\mathbf{x}_{:,1:\frac F2}^{mlp(2)} = \text{MLP}^{(2)}((\mathbf{x}_{:,1:T}^{(scn)})') \in \mathbb{R}^{M\times \frac{F}{2}}.
\end{aligned}
\end{equation}

The output of STCN is the concatenation of the above two parts:
\begin{equation}
    \text{STCN}(\mathbf{x}_{:,1:T}) 
    = \text{concat}(\mathbf{x}_{:,1:\frac F2}^{mlp(1)}, \mathbf{x}_{:, 1:\frac F2}^{mlp(2)}).
\end{equation}


\subsection{Sequence-Guided Mask Attention}

Through the STCN, we obtain the intermediate output $\mathbf{x}_{:,1:F}\in \mathbb{R}^{M\times F}$, and pass it through linear layers to obtain the inputs $\mathbf{Q}, \mathbf{K}, \mathbf{V}$ for the attention function. 

\begin{figure*}[ht]
\centering
\includegraphics[width=1.7\columnwidth]{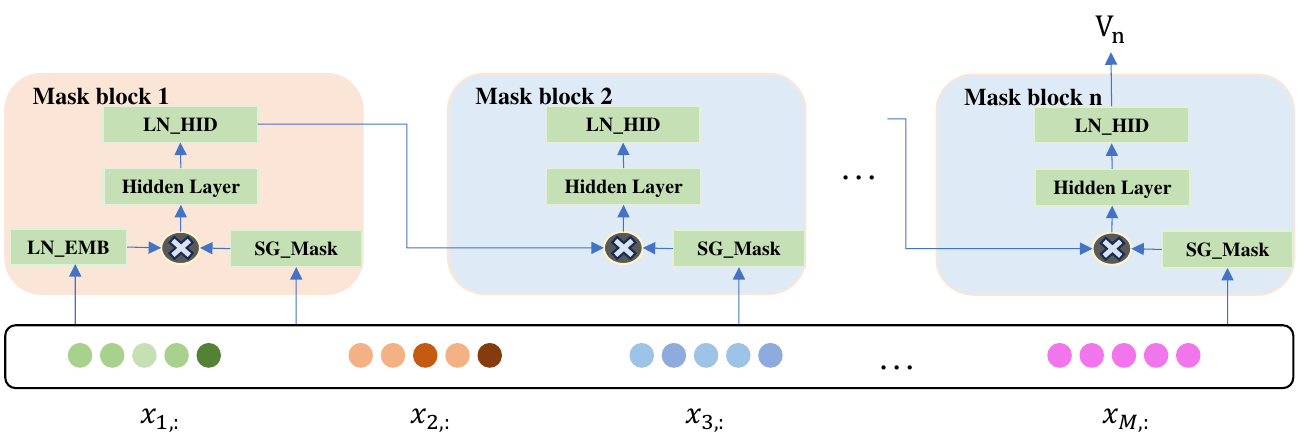}
\caption{Sequence-Guided Mask Attention. This structure extracts contextual features from the embedding inputs ($x_{1,:},x_{2,:},\dots, x_{M,:}$). These features are multiplied by the information directly obtained from the original features through a Sequence-Guided Mask (SG\_Mask) to produce interaction information. The final representation $V_n$, containing global interaction information, is obtained through iterations of $n$ blocks.}
\label{SeqMask}
\end{figure*}

\begin{equation}
\begin{aligned}
    \mathbf{Q} = \mathbf{x}_{:,1:F} \mathbf{W}_\mathbf{Q} \in \mathbb{R}^{M\times d_k}, \\
    \mathbf{K} = \mathbf{x}_{:,1:F} \mathbf{W}_\mathbf{K} \in \mathbb{R}^{M\times d_k}, \\
    \mathbf{V} = \mathbf{x}_{:,1:F} \mathbf{W}_\mathbf{V} \in \mathbb{R}^{M\times d_k}. \\
\end{aligned}
\end{equation}

Drawing on the concept in MaskNet (Wang, She and Zhang 2021), we made adjustments to the attention function by introducing our designed sequence-guided mask function of the $\mathbf{V}$ layer. This approach enables $\mathbf{V}$ to consider global information, while $\mathbf{Q}$ and $\mathbf{k}$ focus more on inter-sequence relationships.

\begin{equation}
    \mathbf{V}_n = \text{SeqMask}(\mathbf{V})
\end{equation}

SeqMask consists of $n$ blocks. For block $i$, the output is $\mathbf{V}_i$, the inputs are the output $V_{i-1}$ of the previous block $i-1$ and $M$ vectors $\mathbf{x}_{:,1:F}$ processed by STCN, i.e, 
$V_i = \text{MaskBlock}_i(V_{i-1}, V_{mask})$. The specific functional form is as follows:
\begin{equation}
\begin{aligned}
    & \mathbf{V}_i = \text{LN}_\text{HID} \left(\mathbf{W}_i * (\mathbf{V}_{i-1} \odot \mathbf{V}_{mask} )\right), \\
    & \text{LN}_\text{HID}(\cdot) = \text{ReLU}(\text{LayerNorm}(\cdot)), \\
    & \mathbf{V}_{mask}  = \text{MLP}^{(4)}(\text{ReLU}(\text{MLP}^{(3)}(\mathbf{V}))).
\end{aligned}
\end{equation}
For block $1$, we use $\text{LN}_\text{EMB}(\mathbf{V})$ to replace the output $V_{i-1}$ of the previous block, 
\begin{equation}
\begin{aligned}
    & \mathbf{V}_1 = \text{LN}_\text{HID} \left(\mathbf{W}_1 * (\text{LN}_\text{EMB}(\mathbf{V}) \odot \mathbf{V}_{mask} )\right), \\
    & \text{LN}_\text{EMB}(\mathbf{V}) = \text{LayerNorm}(\mathbf{V}).
\end{aligned}
\end{equation}
Then, the sequence-guided mask attention can be formulated as follows:
\begin{equation}
    \mathbf{O} = \text{softmax}\left( \frac{\mathbf{Q}\mathbf{K}^T}{\sqrt{d_k}} \right) \mathbf{V}_n \in \mathbb{R}^{M\times d_k}.
\end{equation}

To prevent gradient explosion, the output of the attention mechanism undergoes residual connection and normalization
\begin{equation}
    \mathbf{O}_A = \text{LayerNorm}(\mathbf{O} + \mathbf{x}_{:,1:F}).
\end{equation}

\subsection{FFN}
The remaining parts are the same as in the vanilla Transformer: first through the feed-forward network, followed by the add\&norm operation, to produce the output of the sTransformer block
\begin{equation}
\begin{aligned}
    &\text{FFN}(\mathbf{O}_A) = \text{MLP}^{(6)}(\text{ReLu}(\text{MLP}^{(5)}(\mathbf{O}_A)) ),\\
    &\mathbf{O}_s = \text{LayerNorm}(\text{FFN}(\mathbf{O}_A) + \mathbf{O}_A).
\end{aligned}
\end{equation}

After iterating through multiple sTransformer blocks, the final prediction results are output through projection
\begin{equation}
    \hat{\mathbf{x}}_{:,T+1:T+K} = \text{Projection}(\mathbf{O}_s) 
    = \text{MLP}^{(7)}(\mathbf{O}_s) \in \mathbb{R}^{M\times K}.
\end{equation}

\begin{table*}[ht] 
\centering
\begin{tabular}{c|c|cc|cc|cc|cc|cc|cc}
\toprule[2pt]
\multicolumn{2}{c|}{Methods}  & \multicolumn{2}{c|}{sTransformer} & \multicolumn{2}{c|}{iTransformer} & \multicolumn{2}{c|}{PatchTST} & \multicolumn{2}{c|}{Crossformer}   & \multicolumn{2}{c|}{Informer} & \multicolumn{2}{c|}{DLinear} \\
\midrule
\multicolumn{2}{c|}{Metric}  & MSE & MAE & MSE & MAE& MSE & MAE& MSE & MAE& MSE & MAE& MSE & MAE\\
\midrule[2pt]
\multirow{5}{*}{\rotatebox[origin=c]{90}{ETTh2}}& 96 & \textbf{0.296} & \textbf{0.347} & \underline{0.297} & \underline{0.349} & 0.302 & 0.348 & 0.745 & 0.584  & 3.755 & 1.525& 0.333 & 0.387\\
~ & 192 & \textbf{0.370} & \textbf{0.392} & \underline{0.380} & \underline{0.400} & 0.388 & 0.400 & 0.877 & 0.656  & 5.602 & 1.931 & 0.477 & 0.476\\
~ & 336 & \textbf{0.407} & \textbf{0.426} & \underline{0.428} & \underline{0.432} & 0.426 & 0.433 & 1.043 & 0.731 & 4.721 & 1.835 & 0.594 & 0.541\\
~ & 720 & \textbf{0.414} & \textbf{0.437} & \underline{0.427} & \underline{0.445} & 0.431 & 0.446 & 1.104 & 0.763 & 3.647 & 1.625 & 0.831 & 0.657\\
~ & Avg & \textbf{0.372} & \textbf{0.400} & \underline{0.383} & \underline{0.407} & 0.387 & 0.407 & 0.942 & 0.684 & 4.431 & 1.729 & 0.559 & 0.515\\
\midrule
\multirow{5}{*}{\rotatebox[origin=c]{90}{Electricity}} & 96 & \textbf{0.140} & \textbf{0.238} & \underline{0.148} & \underline{0.240} & 0.195 & 0.285 & 0.219 & 0.314  & 0.274 & 0.368& 0.197 & 0.282 \\
~ & 192 & \textbf{0.158} & \underline{0.254} & \underline{0.162} & \textbf{0.253} & 0.199 & 0.289 & 0.231 & 0.322  & 0.296 & 0.386 & 0.196 & 0.285\\
~ & 336 & \textbf{0.176} & \underline{0.273} & \underline{0.178} & \textbf{0.269} & 0.215 & 0.305 & 0.246 & 0.337  & 0.300 & 0.394 & 0.209 & 0.301\\
~ & 720 & \textbf{0.208} & \textbf{0.300} & \underline{0.225} & \underline{0.317} & 0.256 & 0.337 & 0.280 & 0.363  & 0.373 & 0.439 & 0.245 & 0.333\\
~ & Avg & \textbf{0.171} & \textbf{0.266} & \underline{0.178} & \underline{0.270} & 0.216 & 0.304 & 0.244 & 0.334  & 0.311 & 0.397& 0.212 & 0.300\\ 
\midrule
\multirow{5}{*}{\rotatebox[origin=c]{90}{Traffic}} & 96 & \textbf{0.383} & \textbf{0.266} & \underline{0.395} & \underline{0.268} & 0.544 & 0.359 & 0.522 & 0.290  & 0.719 & 0.391 & 0.650 & 0.396\\
~ & 192 & \textbf{0.403} & \textbf{0.275} & \underline{0.417} & \underline{0.276} & 0.540 & 0.354 & 0.530 & 0.293  & 0.696 & 0.379 & 0.598 & 0.370 \\
~ & 336 & \textbf{0.419} & \textbf{0.282} & \underline{0.433} & \underline{0.283} & 0.551 & 0.358 & 0.558 & 0.305  & 0.777 & 0.420 & 0.605 & 0.373 \\
~ & 720 & \textbf{0.447} & \textbf{0.296} & \underline{0.467} & \underline{0.302} & 0.586 & 0.375 & 0.589 & 0.328  & 0.864 & 0.472 & 0.645 & 0.394 \\
~ & Avg & \textbf{0.413} & \textbf{0.280} & \underline{0.428} & \underline{0.282} & 0.555 & 0.362 & 0.550 & 0.304  & 0.764 & 0.416 & 0.625 & 0.383 \\
\midrule
\multirow{5}{*}{\rotatebox[origin=c]{90}{Weather}} & 96 & \textbf{0.162} & \textbf{0.208} & \underline{0.174} & \underline{0.214} & 0.177 & 0.218 & 0.158 & 0.230 & 0.300 & 0.384 & 0.196 & 0.255\\
~ & 192 & \textbf{0.209} & \textbf{0.251} & \underline{0.221} & \underline{0.254} & 0.225 & 0.259 & 0.206 & 0.277  & 0.598 & 0.544 & 0.261 & 0.237\\
~ & 336 & \textbf{0.266} & \textbf{0.295} & \underline{0.278} & \underline{0.296} & 0.278 & 0.297 & 0.272 & 0.335  & 0.578 & 0.523 & 0.306 & 0.283\\
~ & 720 & \textbf{0.347} & \textbf{0.347} & \underline{0.358} & \underline{0.349} & 0.354 & 0.348 & 0.398 & 0.418  & 1.059 & 0.741 & 0.359 & 0.345\\
~ & Avg & \textbf{0.246} & \textbf{0.275} & \underline{0.258} & \underline{0.279} & 0.259 & 0.281 & 0.259 & 0.315  & 0.634 & 0.548 & 0.287 & 0.265\\
\midrule
\multirow{5}{*}{\rotatebox[origin=c]{90}{Solar-Energy}} & 96 & \textbf{0.196} & \underline{0.238} & \underline{0.203} & \textbf{0.237} & 0.234 & 0.286 & 0.310 & 0.331  & 0.236 & 0.259 & 0.290 & 0.378\\
~ & 192 & \underline{0.229} & \textbf{0.260} & 0.233 & \underline{0.261} & 0.267 & 0.310 & 0.734 & 0.725  & \textbf{0.217} & 0.269 & 0.318 & 0.320\\
~ & 336 & \textbf{0.241} & \textbf{0.271} & \underline{0.248} & \underline{0.273} & 0.29 & 0.315 & 0.750 & 0.735  & 0.249 & 0.283 & 0.330 & 0.353\\
~ & 720 & \underline{0.249} & \underline{0.276} & \underline{0.249} & \textbf{0.275} & 0.289 & 0.317 & 0.769 & 0.765  & \textbf{0.241} & 0.317 & 0.337 & 0.356 \\
~ & Avg & \textbf{0.229} & \textbf{0.261} & \underline{0.233} & \underline{0.262} & 0.270 & 0.307 & 0.641 & 0.639  & 0.235 & 0.280 & 0.319 & 0.330\\
\bottomrule[2pt]
\end{tabular}%
\caption{Performance of different methods on multivariate long-term forecasting tasks with prediction lengths $S\in\{96,192,336,720\}$ and fixed lookback length $T=96$. Five datasets and two evaluation metrics are used here. $Avg$ represents the average value within the dataset. The best values are indicated in \textbf{bold}, and the second best are \underline{underlined}.}
\label{tab_results1}
\end{table*}


\section{Experiment}
\subsection{Datasets}
Public datasets are used to demonstrate the effectiveness of our model. These datasets, often used for comparing time-series forecasting models, include ETT \cite{zhou2021informer}, Electricity, Traffic, Weather used in Autoformer \cite{wu2021autoformer} and Solar-Energy used in LSTNet \cite{lai2018modeling}.

\subsection{Experimental Details}

\subsubsection{Baselines}
We select 9 time-series forecasting models as our benchmark, including iTransformer \cite{liu2023itransformer}, PatchTST \cite{nie2022time}, Crossformer \cite{zhang2023crossformer}, SCINet \cite{liu2022scinet}, TimesNet \cite{wu2022timesnet}, DLinear \cite{zeng2023transformers}, FEDformer \cite{zhou2022fedformer}, Autoformer \cite{wu2021autoformer}, Informer \cite{zhou2021informer}. 


\subsubsection{Main results}
We used commonly used metrics in time-series forecasting, mean squared error (MSE) and mean absolute error (MAE), and adopted MSE as the loss function for training. A lower MSE/MAE means a more accurate forecasting result. Table ???\ref{tab_results1} shows the comparison results. (Due to space constraints, here only list some top-performing models.) Not only did our model outperform linear models, but it also significantly outperformed iTransformer, which was the previous SOTA on five datasets, demonstrating the effectiveness of our approach in capturing sequence correlations.
We achieved the best average MSE/MAE for different lengths across five datasets. Notably, on the ETTh2 and Weather datasets, our model outperformed the existing models across all lengths. Although Crossformer also handles multivariate interactions, sTransformer outperforms it. Our model, on the one hand, utilizes the unique structure of TCN to better extract temporal information, and on the other hand, provides a more effective way to extract multivariate information. 

\begin{table*}[ht]
\centering
\begin{tabular}{c|c|c|cc|cc|cc|cc}
\toprule[2pt]
\multirow{2}{*}{Design}  &\multirow{2}{*}{Temporal} & \multirow{2}{*}{Attention} & \multicolumn{2}{c|}{ETTh2}& \multicolumn{2}{c|}{Electricity}& \multicolumn{2}{c|}{Weather} &\multicolumn{2}{c}{Solar-Energy} \\
\cmidrule{4-11}
~ & ~ & ~ & MSE & MAE & MSE & MAE& MSE & MAE& MSE & MAE \\
\midrule[2pt]
Original & STCN & SeqMask & \textbf{0.372} & 0.400 & 0.171 & 0.266 & \textbf{0.248} & 0.277 & \textbf{0.229} & \textbf{0.261} \\
\midrule
\multirow{2}{*}{Replace} & STCN & Full attention & 0.380 & 0.407 & \textbf{0.169} & \textbf{0.263} & 0.252 & 0.278 & 0.240 & 0.262 \\
\cmidrule{2-11}
~ & FFN & SeqMask & 0.382 & 0.405 & 0.180 & 0.270 & 0.257 & 0.278 & 0.233 & 0.264 \\
\midrule
\multirow{2}{*}{w/o} & STCN & w/o & 0.373 & \textbf{0.398} & 0.174 & 0.268 & 0.249 & \textbf{0.276} & 0.237 & 0.268 \\
\cmidrule{2-11}
~ & w/o & SeqMask & 0.381 & 0.405 & 0.192 & 0.277 & 0.258 & 0.282 & 0.238 & 0.271 \\
\bottomrule[2pt]
\end{tabular}%
\caption{Ablation study on sTransformer. The best values are indicated in \textbf{bold}.}
\label{tab_ablation}
\end{table*}

\begin{figure*}[ht]
\centering
\includegraphics[width=1.75  \columnwidth]{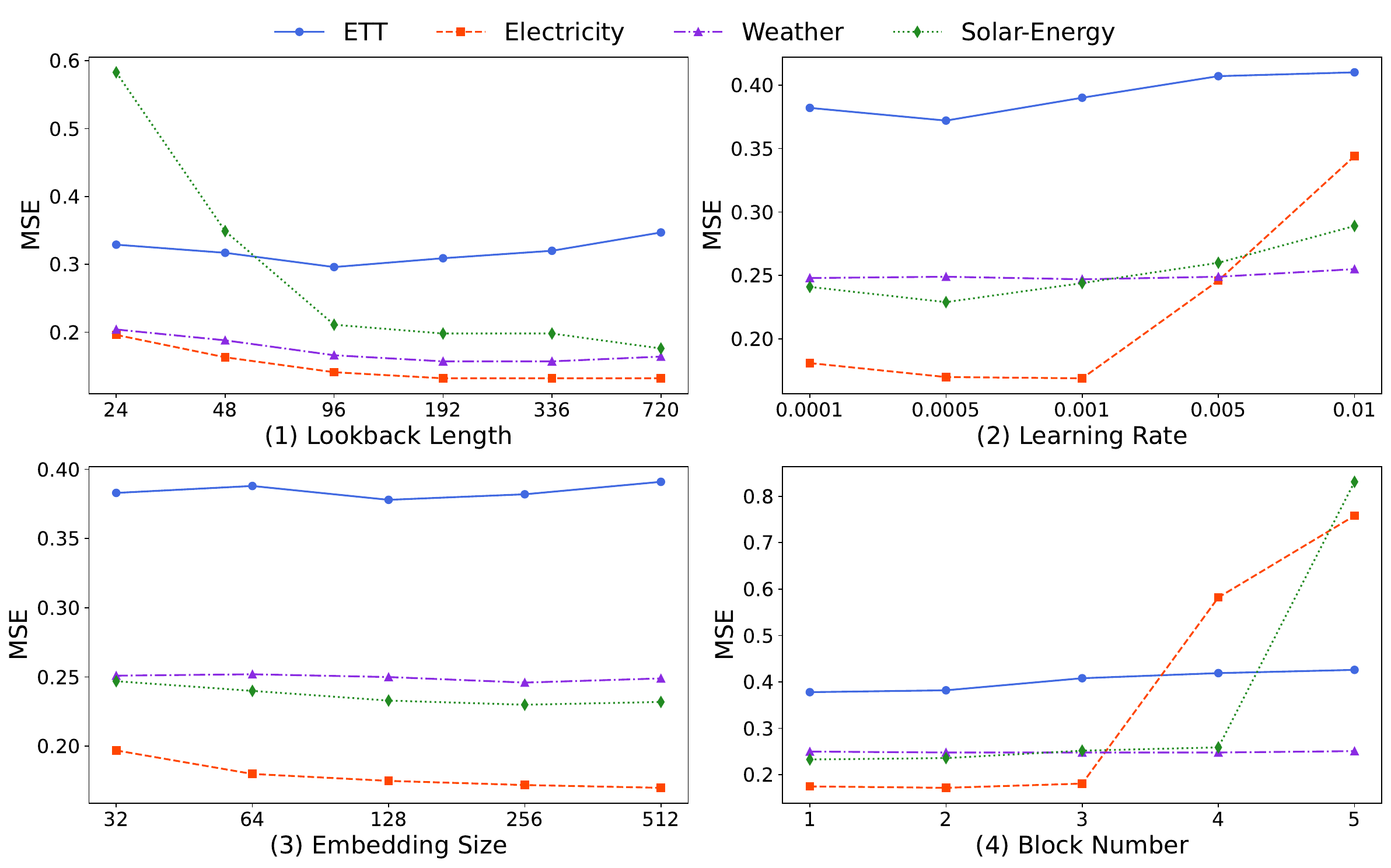}
\caption{Parameter sensitivity. The figure shows the prediction performance of our model with different parameter values on four datasets. The parameters include lookback length, learning rate, embedding size, and block number.}
\label{ParaSens}
\end{figure*}

\begin{table*}[ht] 
\centering
\begin{tabular}{c|c|ccccccccc}
\toprule[2pt]
\multicolumn{2}{c|}{Models}  & sTrans. & TimesNet & N-HiTS & N-BEATS & DLinear &  FED. & Stationay & Auto. & TCN \\
\midrule[2pt]
\multirow{3}{*}{Yearly}& SMAPE & 13.432 & 13.387 & 13.418 & 13.436 & 16.965 & 13.728 & 13.717 & 13.974 & 14.920 \\
~ & MASE & 3.055 &2.996 & 3.045 & 3.043 & 4.283 & 3.048 & 3.078 & 3.134 & 3.364\\
~ & OWA & 0.795 &0.786 & 0.793 & 0.794 & 1.058 & 0.803 & 0.807 & 0.822 & 0.880\\
\midrule
\multirow{3}{*}{Quarterly}& SMAPE & 10.130 &10.100 & 10.202 & 10.124 & 12.145 & 10.792 & 10.958 & 11.338 & 11.122 \\
~ & MASE & 1.190 & 1.182 & 1.194 & 1.169 & 1.520 & 1.283 & 1.325 & 1.365 & 1.360 \\
~ & OWA & 0.894 & 0.890 & 0.899 & 0.886 & 1.106 & 0.958 & 0.981 & 1.012 & 1.001\\
\midrule
\multirow{3}{*}{Monthly}& SMAPE & 12.775 &12.670 & 12.791 & 12.677 & 13.514 & 14.260 & 13.917 & 13.958 & 15.626\\
~ & MASE & 0.949 &0.933 & 0.969 & 0.937 & 1.037 & 1.102 & 1.097 & 1.103 & 1.274\\
~ & OWA & 0.889 &0.878 & 0.899 & 0.880 & 0.956 & 1.012 & 0.998 & 1.002 & 1.141\\
\midrule
\multirow{3}{*}{Others}& SMAPE & 5.075 & 4.891 & 5.061 & 4.925 & 6.709 & 4.954 & 6.302 & 5.485 & 7.186\\
~ & MASE & 3.378 & 3.302 & 3.216 & 3.391 & 4.953 & 3.264 & 4.064 & 3.865 & 4.677\\
~ & OWA & 1.067 & 1.035 & 1.040 & 1.053 & 1.487 & 1.036 & 1.304 & 1.187 & 1.494\\
\midrule
\multirow{3}{*}{\makecell{Weighted\\ Average}}& SMAPE & \textit{11.906} & \textbf{11.829} & 11.927 & \underline{11.851} & 13.639 & 12.840 & 12.780 & 12.909 & 13.961\\
~ & MASE & \textit{1.613} &\textbf{1.585} & \textit{1.613} & \underline{1.599} & 2.095 & 1.701 & 1.756 & 1.771 & 1.945\\
~ & OWA & \textit{0.861} & \textbf{0.851}& \textit{0.861} & \underline{0.855} & 1.051 & 0.918 & 0.930 & 0.939 & 1.023\\
\bottomrule[2pt]
\end{tabular}%
\caption{Performance of different methods in short-term forecasting.  *. means the *former. Some results are based on the data from TimesNet. The best results are indicated in \textbf{bold}, the second are \underline{underlined}, and the third are \textit{italicized}. Our average forecasting performance ranks in the top 3 across metrics SMAPE, MASE and OWA.}
\label{short_results}
\end{table*}

\subsection{Model Analysis}

\subsubsection{Ablation Study}
We conduct additional experiments on datasets with ablation including component replacement (Replace) and component removal (w/o). The results are listed in Table \ref{tab_ablation}. 
We find that the STCN module is the most indispensable component in sTransformer for improving forecasting performance. Both removing it and replacing it with FFN resulted in poorer performance. The SeqMask structure, when replaced with full attention on some datasets, such as Electricity, caused a slight decrease in MSE/MAE. We consider this is due to the specific temporal structure of datasets, where capturing non-essential global information diluted the local information, leading to decreased performance, though the impact was minimal. The ablation study suggest that the use of SeqMask should be considered based on the temporal structure of the data.

\subsubsection{Parameters Sensitivity}
We further analyzed the impact of model parameters on forecasting performance to determine optimal parameters and assess model sensitivity to these parameters (Figure \ref{ParaSens}). Key parameters include lookback length, learning rate, embedding size and block number. 
When the \textbf{lookback length} increases, the MSE of the model gradually decreases (on 3 datasets). A longer lookback window provides more information, thereby improving the forecasting accuracy, which is consistent with the findings mentioned in iTransformer \cite{liu2023itransformer}. For different \textbf{learning rates}, the model performs optimally at 0.0005 and 0.001. Regarding \textbf{embedding size}, larger sizes tend to perform better on datasets with more data, such as Electricity, while smaller datasets like Solar-Energy and Weather show little difference. For the \textbf{block number}, 1-3 blocks are optimal. Increasing the number of blocks to 4-5 may lead to overfitting, resulting in a decline in overall performance.

\subsection{Short-term Forecasting}

Our model achieved state-of-the-art results in long-term forecasting, and we also demonstrated the effectiveness of the model structure in extracting temporal information on short-term forecasting tasks. 
Eight baseline models are include: TimesNet, N-HiTS \cite{challu2022nhitsneuralhierarchicalinterpolation}, N-BEATS \cite{oreshkin2019n}, DLinear, FEDformer, Non-Stationary \cite{liu2022non}, Autoformer and TCN.

\subsubsection{Datasets and Baselines} 
We use the M4 dataset \cite{M4data}, which includes the yearly, quarterly, monthly, weekly, daily and hourly market data. We follow the evaluation framework used in TimesNet \cite{wu2022timesnet}. 

\subsubsection{Main results}
The M4 data is univariate, so it's not possible to perform convolution between sequences. However, we retained the STCN and sequence-guided mask structures. This is equivalent to setting the convolution kernel size between sequences to 1 in the SCN and using only single-variable original inputs in the mask attention. We find that this approach, which can be seen as a self-learning process for the sequence, also provides additional information for forecasting, achieving top 3 performance, close to the performance of TimesNet (Table \ref{short_results}). It demonstrates the generalization ability of our model in prediction tasks.

\begin{table}[ht] 
\centering
\begin{tabular}{c|ccccc}
\toprule[2pt]
Models &sTrans. & Times & iTrans. & Light & DLinear \\
\midrule[2pt]
SMD& \underline{84.09} & \textbf{85.81} & 79.14 & 82.53 & 77.10 \\
\midrule
MSL& 79.18 & \textbf{85.15} & 78.38 & 78.95 & \underline{84.88}\\
\midrule
SWaT& \underline{93.08} & 91.74 & 84.94 & \textbf{93.33} & 87.52\\
\midrule
PSM & 96.25 & \textbf{97.47} & 95.25 & \underline{97.15} & 93.55\\
\midrule
Avg & \underline{88.15} & \textbf{90.04} & 84.42 & 87.99 & 85.76\\
\bottomrule[2pt]
\end{tabular}%
\caption{F1-score (as \%) of different models on anomaly detection task. *. means the *former. $Times$ means TimesNet. $Light$ means LightTS. The best results are indicated in \textbf{bold}, the second are \underline{underlined}.} 
\label{detection_results}
\end{table}

\subsection{Anomaly detection}

\subsubsection{Datasets and Baselines} 
The datasets include SMD \cite{su2019robust}, MSL \cite{hundman2018detecting}, 
SWaT \cite{mathur2016swat} and PSM \cite{abdulaal2021practical}. We also adopt the model evaluation framework from TimesNet, calculating the F1-score for each dataset. Four baseline models are include: TimesNet, iTransformer, LightTS \cite{zhang2022morefastmultivariatetime} and DLinear.

\subsubsection{Main results}
In Table \ref{detection_results}, our model achieves strong performance across all datasets, obtaining the second best performance on average F1-score. 
TimesNet highlights that different tasks require models to have distinct representational abilities, and the representational requirements for time-series forecasting and anomaly detection are similar.
Our results provides additional evidence supporting the viewpoint. 

\section{Conclusion}
In this paper, we study the current state and issues of existing models in time-series forecasting. We propose sTransformer that introduces the STCN module and SeqMask mechanism to capture temporal and multivariate correlations as well as global information representation. Our model combines the strengths of various existing transformer-based models, including strong local and global information representation capabilities and high modular transferability. We conduct experiments on widely used real-world datasets in long-term time-series forecasting and achieve state-of-the-art performance, establishing a new baseline. We conduct additional experiments on short-term forecasting and anomaly detection tasks, achieving top 3 performance, which demonstrate our model's strong information extraction capabilities and generalization ability across tasks for time-series data.

\bibliography{aaai25}

\begin{thebibliography}{35}
\providecommand{\natexlab}[1]{#1}

\bibitem[{Abdulaal, Liu, and Lancewicki(2021)}]{abdulaal2021practical}
Abdulaal, A.; Liu, Z.; and Lancewicki, T. 2021.
\newblock Practical approach to asynchronous multivariate time series anomaly detection and localization.
\newblock In \emph{Proceedings of the 27th ACM SIGKDD conference on knowledge discovery \& data mining}, 2485--2494.

\bibitem[{Bai, Kolter, and Koltun(2018)}]{bai2018empirical}
Bai, S.; Kolter, J.~Z.; and Koltun, V. 2018.
\newblock An empirical evaluation of generic convolutional and recurrent networks for sequence modeling.
\newblock \emph{arXiv preprint arXiv:1803.01271}.

\bibitem[{Challu et~al.(2022)Challu, Olivares, Oreshkin, Garza, Mergenthaler-Canseco, and Dubrawski}]{challu2022nhitsneuralhierarchicalinterpolation}
Challu, C.; Olivares, K.~G.; Oreshkin, B.~N.; Garza, F.; Mergenthaler-Canseco, M.; and Dubrawski, A. 2022.
\newblock N-HiTS: Neural Hierarchical Interpolation for Time Series Forecasting.
\newblock arXiv:2201.12886.

\bibitem[{Das et~al.(2023)Das, Kong, Leach, Mathur, Sen, and Yu}]{das2023long}
Das, A.; Kong, W.; Leach, A.; Mathur, S.; Sen, R.; and Yu, R. 2023.
\newblock Long-term forecasting with tide: Time-series dense encoder.
\newblock \emph{arXiv preprint arXiv:2304.08424}.

\bibitem[{Dosovitskiy et~al.(2020)Dosovitskiy, Beyer, Kolesnikov, Weissenborn, Zhai, Unterthiner, Dehghani, Minderer, Heigold, Gelly et~al.}]{dosovitskiy2020image}
Dosovitskiy, A.; Beyer, L.; Kolesnikov, A.; Weissenborn, D.; Zhai, X.; Unterthiner, T.; Dehghani, M.; Minderer, M.; Heigold, G.; Gelly, S.; et~al. 2020.
\newblock An image is worth 16x16 words: Transformers for image recognition at scale.
\newblock \emph{arXiv preprint arXiv:2010.11929}.

\bibitem[{Franceschi, Dieuleveut, and Jaggi(2019)}]{franceschi2019unsupervised}
Franceschi, J.-Y.; Dieuleveut, A.; and Jaggi, M. 2019.
\newblock Unsupervised scalable representation learning for multivariate time series.
\newblock \emph{Advances in neural information processing systems}, 32.

\bibitem[{Gillioz et~al.(2020)Gillioz, Casas, Mugellini, and Abou~Khaled}]{gillioz2020overview}
Gillioz, A.; Casas, J.; Mugellini, E.; and Abou~Khaled, O. 2020.
\newblock Overview of the Transformer-based Models for NLP Tasks.
\newblock In \emph{2020 15th Conference on computer science and information systems (FedCSIS)}, 179--183. IEEE.

\bibitem[{Huang et~al.(2020)Huang, Hu, Yeung, and Chen}]{huang2020conv}
Huang, W.; Hu, W.; Yeung, Y.~T.; and Chen, X. 2020.
\newblock Conv-transformer transducer: Low latency, low frame rate, streamable end-to-end speech recognition.
\newblock \emph{arXiv preprint arXiv:2008.05750}.

\bibitem[{Hundman et~al.(2018)Hundman, Constantinou, Laporte, Colwell, and Soderstrom}]{hundman2018detecting}
Hundman, K.; Constantinou, V.; Laporte, C.; Colwell, I.; and Soderstrom, T. 2018.
\newblock Detecting spacecraft anomalies using lstms and nonparametric dynamic thresholding.
\newblock In \emph{Proceedings of the 24th ACM SIGKDD international conference on knowledge discovery \& data mining}, 387--395.

\bibitem[{Kalyan, Rajasekharan, and Sangeetha(2021)}]{kalyan2021ammus}
Kalyan, K.~S.; Rajasekharan, A.; and Sangeetha, S. 2021.
\newblock Ammus: A survey of transformer-based pretrained models in natural language processing.
\newblock \emph{arXiv preprint arXiv:2108.05542}.

\bibitem[{Karita et~al.(2019)Karita, Chen, Hayashi, Hori, Inaguma, Jiang, Someki, Soplin, Yamamoto, Wang et~al.}]{karita2019comparative}
Karita, S.; Chen, N.; Hayashi, T.; Hori, T.; Inaguma, H.; Jiang, Z.; Someki, M.; Soplin, N. E.~Y.; Yamamoto, R.; Wang, X.; et~al. 2019.
\newblock A comparative study on transformer vs rnn in speech applications.
\newblock In \emph{2019 IEEE automatic speech recognition and understanding workshop (ASRU)}, 449--456. IEEE.

\bibitem[{Lai et~al.(2018)Lai, Chang, Yang, and Liu}]{lai2018modeling}
Lai, G.; Chang, W.-C.; Yang, Y.; and Liu, H. 2018.
\newblock Modeling long-and short-term temporal patterns with deep neural networks.
\newblock In \emph{The 41st international ACM SIGIR conference on research \& development in information retrieval}, 95--104.

\bibitem[{Li et~al.(2019)Li, Jin, Xuan, Zhou, Chen, Wang, and Yan}]{li2019enhancing}
Li, S.; Jin, X.; Xuan, Y.; Zhou, X.; Chen, W.; Wang, Y.-X.; and Yan, X. 2019.
\newblock Enhancing the locality and breaking the memory bottleneck of transformer on time series forecasting.
\newblock \emph{Advances in neural information processing systems}, 32.

\bibitem[{Lim et~al.(2021)Lim, Ar{\i}k, Loeff, and Pfister}]{lim2021temporal}
Lim, B.; Ar{\i}k, S.~{\"O}.; Loeff, N.; and Pfister, T. 2021.
\newblock Temporal fusion transformers for interpretable multi-horizon time series forecasting.
\newblock \emph{International Journal of Forecasting}, 37(4): 1748--1764.

\bibitem[{Liu et~al.(2022{\natexlab{a}})Liu, Zeng, Chen, Xu, Lai, Ma, and Xu}]{liu2022scinet}
Liu, M.; Zeng, A.; Chen, M.; Xu, Z.; Lai, Q.; Ma, L.; and Xu, Q. 2022{\natexlab{a}}.
\newblock Scinet: Time series modeling and forecasting with sample convolution and interaction.
\newblock \emph{Advances in Neural Information Processing Systems}, 35: 5816--5828.

\bibitem[{Liu et~al.(2023{\natexlab{a}})Liu, Hu, Zhang, Wu, Wang, Ma, and Long}]{liu2023itransformer}
Liu, Y.; Hu, T.; Zhang, H.; Wu, H.; Wang, S.; Ma, L.; and Long, M. 2023{\natexlab{a}}.
\newblock itransformer: Inverted transformers are effective for time series forecasting.
\newblock \emph{arXiv preprint arXiv:2310.06625}.

\bibitem[{Liu et~al.(2022{\natexlab{b}})Liu, Wu, Wang, and Long}]{liu2022non}
Liu, Y.; Wu, H.; Wang, J.; and Long, M. 2022{\natexlab{b}}.
\newblock Non-stationary transformers: Exploring the stationarity in time series forecasting.
\newblock \emph{Advances in Neural Information Processing Systems}, 35: 9881--9893.

\bibitem[{Liu et~al.(2023{\natexlab{b}})Liu, Zhang, Wang, Hou, Yuan, Tian, Zhang, Shi, Fan, and He}]{liu2023survey}
Liu, Y.; Zhang, Y.; Wang, Y.; Hou, F.; Yuan, J.; Tian, J.; Zhang, Y.; Shi, Z.; Fan, J.; and He, Z. 2023{\natexlab{b}}.
\newblock A survey of visual transformers.
\newblock \emph{IEEE Transactions on Neural Networks and Learning Systems}.

\bibitem[{Makridakis.(2018)}]{M4data}
Makridakis., S. 2018.
\newblock M4 dataset.
\newblock \url{https://github.com/M4Competition/ M4-methods/tree/master/Dataset}.

\bibitem[{Mathur and Tippenhauer(2016)}]{mathur2016swat}
Mathur, A.~P.; and Tippenhauer, N.~O. 2016.
\newblock SWaT: A water treatment testbed for research and training on ICS security.
\newblock In \emph{2016 international workshop on cyber-physical systems for smart water networks (CySWater)}, 31--36. IEEE.

\bibitem[{Nie et~al.(2022)Nie, Nguyen, Sinthong, and Kalagnanam}]{nie2022time}
Nie, Y.; Nguyen, N.~H.; Sinthong, P.; and Kalagnanam, J. 2022.
\newblock A time series is worth 64 words: Long-term forecasting with transformers.
\newblock \emph{arXiv preprint arXiv:2211.14730}.

\bibitem[{Oreshkin et~al.(2019)Oreshkin, Carpov, Chapados, and Bengio}]{oreshkin2019n}
Oreshkin, B.~N.; Carpov, D.; Chapados, N.; and Bengio, Y. 2019.
\newblock N-BEATS: Neural basis expansion analysis for interpretable time series forecasting.
\newblock \emph{arXiv preprint arXiv:1905.10437}.

\bibitem[{Su et~al.(2019)Su, Zhao, Niu, Liu, Sun, and Pei}]{su2019robust}
Su, Y.; Zhao, Y.; Niu, C.; Liu, R.; Sun, W.; and Pei, D. 2019.
\newblock Robust anomaly detection for multivariate time series through stochastic recurrent neural network.
\newblock In \emph{Proceedings of the 25th ACM SIGKDD international conference on knowledge discovery \& data mining}, 2828--2837.

\bibitem[{Vaswani et~al.(2017)Vaswani, Shazeer, Parmar, Uszkoreit, Jones, Gomez, Kaiser, and Polosukhin}]{vaswani2017attention}
Vaswani, A.; Shazeer, N.; Parmar, N.; Uszkoreit, J.; Jones, L.; Gomez, A.~N.; Kaiser, {\L}.; and Polosukhin, I. 2017.
\newblock Attention is all you need.
\newblock \emph{Advances in neural information processing systems}, 30.

\bibitem[{Wang et~al.(2022)Wang, Wang, Li, Gu, Lu, Zhang, and Gu}]{wang2022enhancing}
Wang, F.; Wang, Y.; Li, D.; Gu, H.; Lu, T.; Zhang, P.; and Gu, N. 2022.
\newblock Enhancing CTR prediction with context-aware feature representation learning.
\newblock In \emph{Proceedings of the 45th International ACM SIGIR Conference on Research and Development in Information Retrieval}, 343--352.

\bibitem[{Wang, She, and Zhang(2021)}]{wang2021masknet}
Wang, Z.; She, Q.; and Zhang, J. 2021.
\newblock Masknet: Introducing feature-wise multiplication to CTR ranking models by instance-guided mask.
\newblock \emph{arXiv preprint arXiv:2102.07619}.

\bibitem[{Wen et~al.(2022)Wen, Zhou, Zhang, Chen, Ma, Yan, and Sun}]{wen2022transformers}
Wen, Q.; Zhou, T.; Zhang, C.; Chen, W.; Ma, Z.; Yan, J.; and Sun, L. 2022.
\newblock Transformers in time series: A survey.
\newblock \emph{arXiv preprint arXiv:2202.07125}.

\bibitem[{Wu et~al.(2022)Wu, Hu, Liu, Zhou, Wang, and Long}]{wu2022timesnet}
Wu, H.; Hu, T.; Liu, Y.; Zhou, H.; Wang, J.; and Long, M. 2022.
\newblock Timesnet: Temporal 2d-variation modeling for general time series analysis.
\newblock \emph{arXiv preprint arXiv:2210.02186}.

\bibitem[{Wu et~al.(2021{\natexlab{a}})Wu, Xiao, Codella, Liu, Dai, Yuan, and Zhang}]{wu2021cvt}
Wu, H.; Xiao, B.; Codella, N.; Liu, M.; Dai, X.; Yuan, L.; and Zhang, L. 2021{\natexlab{a}}.
\newblock Cvt: Introducing convolutions to vision transformers.
\newblock In \emph{Proceedings of the IEEE/CVF international conference on computer vision}, 22--31.

\bibitem[{Wu et~al.(2021{\natexlab{b}})Wu, Xu, Wang, and Long}]{wu2021autoformer}
Wu, H.; Xu, J.; Wang, J.; and Long, M. 2021{\natexlab{b}}.
\newblock Autoformer: Decomposition transformers with auto-correlation for long-term series forecasting.
\newblock \emph{Advances in neural information processing systems}, 34: 22419--22430.

\bibitem[{Zeng et~al.(2023)Zeng, Chen, Zhang, and Xu}]{zeng2023transformers}
Zeng, A.; Chen, M.; Zhang, L.; and Xu, Q. 2023.
\newblock Are transformers effective for time series forecasting?
\newblock In \emph{Proceedings of the AAAI conference on artificial intelligence}, volume~37, 11121--11128.

\bibitem[{Zhang et~al.(2022)Zhang, Zhang, Cao, Bian, Yi, Zheng, and Li}]{zhang2022morefastmultivariatetime}
Zhang, T.; Zhang, Y.; Cao, W.; Bian, J.; Yi, X.; Zheng, S.; and Li, J. 2022.
\newblock Less Is More: Fast Multivariate Time Series Forecasting with Light Sampling-oriented MLP Structures.
\newblock arXiv:2207.01186.

\bibitem[{Zhang and Yan(2023)}]{zhang2023crossformer}
Zhang, Y.; and Yan, J. 2023.
\newblock Crossformer: Transformer utilizing cross-dimension dependency for multivariate time series forecasting.
\newblock In \emph{The eleventh international conference on learning representations}.

\bibitem[{Zhou et~al.(2021)Zhou, Zhang, Peng, Zhang, Li, Xiong, and Zhang}]{zhou2021informer}
Zhou, H.; Zhang, S.; Peng, J.; Zhang, S.; Li, J.; Xiong, H.; and Zhang, W. 2021.
\newblock Informer: Beyond efficient transformer for long sequence time-series forecasting.
\newblock In \emph{Proceedings of the AAAI conference on artificial intelligence}, volume~35, 11106--11115.

\bibitem[{Zhou et~al.(2022)Zhou, Ma, Wen, Wang, Sun, and Jin}]{zhou2022fedformer}
Zhou, T.; Ma, Z.; Wen, Q.; Wang, X.; Sun, L.; and Jin, R. 2022.
\newblock Fedformer: Frequency enhanced decomposed transformer for long-term series forecasting.
\newblock In \emph{International conference on machine learning}, 27268--27286. PMLR.

\end{thebibliography}

\end{document}